\documentclass{article}
\usepackage{spconf,amsmath,graphicx}
\usepackage[toc,page]{appendix}
\usepackage{multirow}
\usepackage{booktabs}
\usepackage[final]{pdfpages}
\usepackage{graphicx}

\newcommand{\Inp}{\mathcal{X}}
\newcommand{\Feat}{\mathcal{Z}}
\newcommand{\QFeat}{\mathcal{Q}}
\newcommand{\Context}{\mathcal{C}}

\newcommand{\ze}{\mathbf{z}}
\newcommand{\zq}{\mathbf{q}}

\newcommand{\cc}{\mathbf{c}}

\title{Simple and Effective Zero-shot Cross-lingual Phoneme Recognition}
\name{Qiantong Xu, Alexei Baevski, Michael Auli}
\address{Facebook AI Research}
\begin{document}
%
\maketitle
\begin{abstract}
Recent progress in self-training, self-supervised pretraining and unsupervised learning enabled well performing speech recognition systems without any labeled data.
However, in many cases there is labeled data available for related languages which is not utilized by these methods.
This paper extends previous work on zero-shot cross-lingual transfer learning by fine-tuning a multilingually pretrained wav2vec 2.0 model to transcribe unseen languages.
This is done by mapping phonemes of the training languages to the target language using articulatory features.
Experiments show that this simple method significantly outperforms prior work which introduced task-specific architectures and used only part of a monolingually pretrained model.
\end{abstract}
\section{Introduction}
\label{sec:intro}


There is a large number of languages spoken around the world of which only a small fraction is served by speech technology. 
A large barrier to making speech technology more accessible is the requirement for large amounts of transcribed speech audio by current models which is simply not available for the vast majority of languages.
Speech recognition accuracy has been steadily improving by recent advances in supervised multilingual modeling~\cite{dalmia2018sequence, pratap2020massively}, self-supervised learning~\cite{oord2018cpc,chung2018speech2vec,chung2019unsupervised,baevski2020wav2vec, hsu2021hubert}, and semi-supervised learning~\cite{synnaeve2019end, xu2020iterative, likhomanenko2020slimipl, xu2021self, park2020improved}, particularly for low-resource languages.
This recently led to good speech recognition performance in settings where no labeled data exists at all~\cite{liu2018completely,chen2019completely,baevski2021unsupervised}.
One downside of these approaches is that they require training a separate unsupervised model for each language while ignoring the presence of labeled data in related languages.

Zero-shot transfer learning addresses this by training a single multilingual model on the labeled data of several languages to enable zero-shot transcription of unseen languages~\cite{gaozero, li2020universal, jacobs2021multilingual, yan2021differentiable, li2020universal, li2020towards}.
Models usually have a common encoder that extracts acoustic information from speech audio and then predict either a shared phoneme vocabulary~\cite{li2020universal, gaozero} or language-specific phonemes~\cite{dalmia2018sequence, li2020towards, winata2020adapt}. 
The former requires either phonological units that are agnostic to any particular language such as articulatory features~\cite{li2020towards} or global phones \cite{schultz2002globalphone, li2020universal}. 

In this paper, we study a simple zero-shot transfer learning approach which builds a global phoneme recognizer by simply considering all possible phonemes of the training languages and then decodes the model with a language model to generate the final phoneme sequence. 
The lexicon is built from articulatory features to map the phonemes between the training and target vocabulary. 
Our method makes no assumption about the relation of training and testing languages, including attributes like phoneme distribution or coverage.
We extend prior work by using unsupervised cross-lingually pretrained representations estimated on 53 languages~\cite{conneau2020unsupervised} instead of monolingually trained representations~\cite{gaozero} and our approach also uses the full pretrained model instead of only the feature-extractor~\cite{gaozero}.

We conduct experiments on 42 languages of CommonVoice~\cite{ardila2019common}, 19 languages of BABEL~\cite{gales2014speech} and six languages of MLS~\cite{pratap2020mls}. Results show significant improvements on unseen languages over the approach of~\cite{gaozero} and cross-lingual pretrained representations are more effective.
Finally, zero-shot transfer learning performs comparably to unsupervised approaches with the benefit of being able to transcribe multiple unseen languages using a single model.

\section{Approach}
\label{sec:approach}

Our approach entails the use of self-supervised representations trained on data in many languages (\cite{conneau2020unsupervised}, \textsection\ref{sec:w2v}).
Next we simultaneously fine-tune the model to perform phoneme recognition on data in multiple training languages.
At inference time, we test the fine-tuned model on all unseen languages using a mapping of the phonemes from the training vocabulary to the ones in the target languages (\textsection\ref{sec:mapping}).

\subsection{Self-supervised Model Training}
\label{sec:w2v}

We use XLSR-53, a wav2vec 2.0 model pretrained on data in 53 languages~(\cite{conneau2020unsupervised,baevski2020wav2vec}). 
This model contains a convolutional feature encoder $f: \Inp \mapsto \Feat$ to map raw audio~$\Inp$ to latent speech representations $\ze_1, \dots, \ze_T$ which are input to a Transformer $g: \Feat \mapsto \Context$ to output context representations $\cc_1, \dots, \cc_T$~\cite{devlin2018bert,baevski2019vqwav2vec}.
Each $\ze_t$ represents about 25ms of audio strided by 20ms and the Transformer architecture follows BERT~\cite{vaswani2017transformer,devlin2018bert}.
During training, feature encoder representations are discretized to $\zq_1, \dots, \zq_T$ with a quantization module $\Feat \mapsto \QFeat$ to represent the targets in the objective.
The quantization module uses a Gumbel softmax to choose entries form the codebooks and the chosen entries are concatenated to obtain $\zq$~\cite{jegou2011ieee,jang2016gumbel,baevski2019vqwav2vec}.
The model is trained by solving a contrastive task over masked feature encoder outputs.
At training time, spans of ten time steps with random starting indices are masked.
The objective requires identifying the true quantized latent $\zq_t$ for a masked time-step within a set of $K=100$ distractors $\mathbf{Q}_t$ sampled from other masked time steps.


\subsection{Phoneme Mapping}
\label{sec:mapping}

We use phonemes as modeling units and in particular, the symbols of the standard International Phonetic Alphabet (IPA). 
However, the vocabulary estimated from the training languages may not cover the full vocabulary of the target languages which results in out-of-vocabulary (OOV) phonemes at test time.
We address this by mapping between the training and target vocabularies based on articulatory/phonological features~\cite{Mortensen-et-al:2016}. 
Articulatory feature is a set of global attributes to describe any sound or phone. 
There are four groups of attributes: major class (syllabic, vocalic, approximant, sonorant), manner (continuant, lateral, nasal, strident), place (labial, coronal, dorsal, pharyngeal), and laryngeal (voiced, aspirated, glottalized). 
Each attribute can be either positive or negative.

We compute the distance between each pair of phonemes using the hamming edit distance between the articulatory feature vectors\footnote{https://github.com/dmort27/panphon. In this repository, each feature articulatory vector contains 21 attributes}, and then generate two types of simple many-to-one mapping lexicons:
\begin{itemize}
\item \textbf{tr2tgt lexicon} maps each phoneme in the training vocabulary to its closest one in the target vocabulary. Then for the remaining uncovered phonemes in the target vocabulary, it maps the closest ones in the training vocabulary to them.
\item \textbf{tgt2tr lexicon} that maps for each phoneme in the target vocabulary, the phonemes in the training vocabulary that have 0 distance to it.  
\end{itemize}
We compare both below (\textsection\ref{sec:exp_lexicons}) and use tr2tgt unless otherwise mentioned.

\section{Experimental setup}
\label{sec:expsetup}

\begin{table}[t]
\caption{Splits of CommonVoice (CV) and BABEL (BB). The 6 BABEL languages of \cite{gaozero} are bolded. 
\label{tab:data}}
\centering

\begin{tabular}{@{}ll@{}}
\toprule
\textbf{Split}         & \textbf{Languages}                                   \\
\midrule
\multicolumn{2}{l}{CommonVoice (CV)} \\
\multirow{7}{*}{train} & Esperanto(eo), Lithuanian(lt), Welsh(cy), Tamil(ta),    \\
                       & Swedish(sv-SE), German(de), English(en), Oriya(or),  \\
                       & Hindi(hi), Persian(fa), Japanese(ja), Assamese(as),       \\
                       & Indonesian(id), Catalan(ca), Spanish(es), French(fr),   \\
                       & Portuguese(pt), Arabic(ar), Chinese(zh-CN),            \\
                       & Chinese(zh-TW), Turkish(tr), Estonian(et),             \\
                       & Hungarian(hu), Russian(ru), Czech(cs)                  \\
\cmidrule(lr){2-2}
dev                    & Italian(it)                                          \\
\cmidrule(lr){2-2}
\multirow{4}{*}{test}  & Basque(eu), Interlingua(ia), Latvian(lv), Georgian(ka), \\
                       & Irish(ga-IE), Dutch(nl), Greek(el), Punjabi(pa-IN),     \\
                       & Romanian(ro), Maltese(mt), Chinese(zh-HK), Tatar(tt),   \\
                       & Finnish(fi), Slovenian(sl), Polish(pl), Kirghiz(ky)     \\
\midrule
\multicolumn{2}{l}{BABEL (BB)} \\
\multirow{5}{*}{train} & \textbf{Amharic(am)}, \textbf{Bengali(bn)}, Cebuano(ceb), Igbo(ig),  \\
                       & Haitian(ht), \textbf{Javanese(jv)}, Mongolian(mn), Swahili(sw),  \\
                       & Tamil(ta), \textbf{Vietnamese(vi)}, Assamese(as), Dholuo(luo),   \\
                       & Guarani(gn), Kazakh(kk), Pashto(ps), \textbf{Georgian(ka)},         \\
                       & Tagalog(tl), Telugu(te), Turkish(tr), \textbf{Zulu(zu)}          \\
\cmidrule(lr){2-2}
dev                    & CV-Italian(it)                                       \\
\cmidrule(lr){2-2}
test                   & Cantonese(yue), Lao(lo)     \\
\bottomrule
\end{tabular}
\end{table}

\subsection{Datasets}
We consider three multilingual corpora and a variety of languages to evaluate our approach. All the audios are up-/down-sampled to 16kHz.

\textbf{Multilingual LibriSpeech (MLS)} is a large corpus of read audiobooks from Librivox and we experiment with the same six languages as~\cite{baevski2021unsupervised}: Dutch (du), French (fr), German (de), Italian (it), Portuguese (pt), Spanish (es). We use the same split as~\cite{baevski2021unsupervised} for validation and test.

\textbf{CommonVoice (CV)} is a multilingual corpus of read speech comprising more than two thousand hours of speech data in 76 languages~\cite{ardila2019common}. 
We use the December 2020 release (v6.1) for training and fine-tuning models. 
We select 42 languages in total that are supported by our phonemizer (see \textsection\ref{sec:preprocess}) as well as their official train, dev and test splits.
Italian (it) serves as validation language for development, for training we use a total of 26 languages and the remaining 13 languages are for testing (Table~\ref{tab:data}). 
For each language in the test set, we also make sure that there is at least one language that belongs to the same language family as in the training set. 
Compared to other datasets such as BABEL or MLS, CommonVoice is well suited for zero-shot transfer learning, since it covers a larger number of languages.

\textbf{BABEL} is a multilingual corpus of conversational telephone speech from IARPA, which includes Asian and African language \cite{gales2014speech}. 
We include 21 languages from it (Table~\ref{tab:data}). 
We include Cantonese and Lao in the test set to compare with \cite{gaozero} and the remaining 19 languages in the training set. Italian serves for validation.

\subsection{Pre-processing and Phonemization}
\label{sec:preprocess}
We first normalize all transcriptions for CommonVoice and BABEL by removing punctuation and rare characters. 
Rare characters are usually numbers or characters from other languages. 
We then obtain the phonemic annotations from the word transcriptions using ESpeak\footnote{https://github.com/espeak-ng/espeak-ng}, as well as \cite{hasegawa2020grapheme} based on Phonetisaurus\footnote{https://github.com/AdolfVonKleist/Phonetisaurus} to compare with \cite{gaozero}. 
Specifically, we use Espeak on MLS, Phonetisaurus on BABEL. 

\subsection{Model Training}
Models are implemented in fairseq \cite{ott2019fairseq} and we use the pretrained XLSR-53 model~\cite{conneau2020unsupervised} which has 24 Transformer blocks, model dimension 1024, inner dimension 4096 and 16 attention heads. 
It is pretrained on the joint training set of MLS, CommonVoice and BABEL, which consists of about 56K hours of speech data.

To fine-tune the model we add a classifier representing the joint vocabulary of the training languages on top of the model and train on the labeled data with a Connectionist
Temporal Classification (CTC) loss \cite{graves2006connectionist}. Weights of the feature encoder are not updated at fine-tuning time, while the Transformer weights are finetuned after 10k updates. We determine the best transformer final dropout in [0, 0.3], learning rates setting in [5e-6, 5e-4]. The learning rate schedule has three phases: warm up for the first 10\% of updates, keep constant for 40\% and then linearly decay for the remainder. The models were finetuned for 25k updates on 4 GPUs. The best checkpoints are selected by the validation error on the validations set for BABEL and CommonVoice; while for MLS, it is selected using the unsupervised cross validation metric of \cite{baevski2021unsupervised} to enable a direct comparison.


\subsection{Decoding}

The wav2letter beam-search decoder~\cite{pratap2019wav2letter++} is used to generate the final transcriptions with the lexicon and an external 6-gram language model trained on the phoneme annotations of the labeled training data.
Beam size is set to 50 in all the inference experiments. 
The lexicons mentioned above limits the search space to only the valid phones in the training vocabulary and ensures the decoder predicts only phones in the target dictionary.


\section{Results}
\label{sec:res}

\subsection{Comparison with unsupervised method}
\begin{table}[t]
\begin{center}
    
\caption{Unsupervised ASR (w2v-U) vs. zero-shot ASR (This work). Results are in terms of phoneme error rate (PER) on MLS.
\label{tab:mls_res}
}
\setlength{\tabcolsep}{5pt}
\begin{tabular}{@{}lcccccc|c@{}}
\toprule
            & de             & nl    & fr             & es             & it             & pt    & Avg \\ 
\midrule
w2v-U \cite{baevski2021unsupervised}      & 21.6          & 25.0 & 27.7          & 20.2          & 31.2          & 36.0 & 27.0 \\
+ n-gram LM                                           & 16.2          & 17.8 & 26.5          & 18.1          & 28.6          & 30.6 & 23.0 \\
\midrule
This work   & 23.8          & 38.0 & 31.0          & 28.7          & 33.5          & 45.0 & 33.3 \\
+ n-gram LM                                           & 14.8 & 26.0 & 26.4 & 12.3 & 21.7 & 36.5 & 22.9 \\ 
\bottomrule
\end{tabular}
\end{center}
\end{table}

In our first experiment, we compare zero-shot transfer learning to wav2vec-U~\cite{baevski2021unsupervised}, both of which use the same pretrained representations (XLSR-53).
We use 10 hours of labeled data for each MLS language as prepared in  \cite{conneau2020unsupervised} and measure the performance when fine-tuning XLSR-53 on five of the six languages and then evaluate on the held-out language.
Table~\ref{tab:mls_res} shows that the performance of zero-shot transfer learning is on par to wav2vec-U~\cite{baevski2021unsupervised} while using a simpler training and inference pipeline.

\subsection{Comparison to other zero-shot work}
\begin{table}[t]
\begin{center}
\caption{Comparison to prior zero-shot work~\cite{gaozero} in terms of phonetic token error rate (PTER) on the test sets of a subset of BABEL languages. Cantonese and Lao are the unseen languages. Models are trained on 6 or 19 languages of BABEL (BB-6/19), 21 languages of CommonVoice (CV-21), Globalphone (GP) and the Spoken Dutch Corpus (CGN).\label{tab:gao_res}
}
\setlength{\tabcolsep}{4.5pt}
\begin{tabular}{@{}lc|c|cc@{}}
\toprule
& & Gao et al. \cite{gaozero} & \multicolumn{2}{|c}{This work} \\
\midrule
& BB Data          & BB-6 & BB-6 & BB-19 \\
& Other Data       & CGN+GP & - & CV-21 \\
\midrule 
& \# hours / lang &        all              & all & 10                        \\
& \# hours total               & 1,492                & 317            & 298                       \\
\midrule
\multirow{6}{*}{Supervised } & Bengali                   & 38.2                 & \textbf{36.1}    & 40.7                      \\
& Vietnamese                & \textbf{32.0}        & 40.7             & 63.3                      \\
& Zulu                      & 35.2                 & \textbf{34.6}    & 44.1                      \\
& Amharic                   & 38.0                 & \textbf{35.5}    & 42.8                      \\
& Javanese                  & 44.2                 & \textbf{40.2}    & 49.1                      \\
& Georgian                  & 38.6                 & \textbf{27.6}    & 43.2                      \\
\midrule
\multirow{2}{*}{Zero-shot} & Cantonese                 & 73.1                 & 73.6             & \textbf{63.6}             \\
& Lao                       & 69.3                 & 70.3             & \textbf{63.7}    \\ 
\bottomrule         
\end{tabular}

\end{center}
\end{table}

Next, we compare performance to the zero-shot transfer learning approach of \cite{gaozero} which used only the feature extractor of a wav2vec 2.0 model trained on English.
The training data on CommonVoice and BABEL is prepared in the same way as \cite{gaozero} and we report the same phonetic token error rate (PTER) metric, in which each IPA token is treated as separate suprasegmentals (such as long vowels, and primary stress symbol), tones, diphthongs and affricates.

Table~\ref{tab:gao_res} shows that finetuning on only 6 languages of BABEL (BB-6) with our method can outperform \cite{gaozero} on the supervised languages while using 317 hours of labeled data compared to nearly 1.5K hours.
This shows that using the full pretraining model is beneficial.
Our approach can outperform~\cite{gaozero} on the zero-shot directions when we add CommonVoice data while restricting the amount of labeled data to 10 hours for each language. 
This results in fewer than 300 hours of labeled data since some languages do not even have 10 hours.

\subsection{Ablations}
\begin{table}[t]
\caption{Effect of no pretraining, monolingual pre-training (w2v LV-60K) and multilingual pretraining (XLSR-53) in terms of PER on the test languages of CommonVoice. \label{tab:cv_res_abl1}}
\begin{center}
\begin{tabular}{@{}ccccc@{}}
\toprule
  & \multicolumn{2}{c}{No pretrain} & w2v LV-60K & XLSR-53   \\
 \# hours / lang & 10 & 200 & 10 & 10 \\
 \# hours total & 149 & 1156 & 149 & 149 \\
\midrule
it       & 47.5   & 41.8      & 16.9          & 13.9   \\
eu       & 45.6   & 32.1      & 16.3          & 13.7             \\
ia       & 27.8      & 23.0               & 6.7     & 6.1      \\
lv       & 59.8      & 56.5               & 33.5    & 32.3    \\
ka       & 56.1      & 48.9               & 24.0    & 23.8    \\
nl       & 56.8      & 56.1               & 30.5    & 19.8    \\
el       & 40.6      & 33.7               & 10.7    & 10.4             \\
ro       & 34.7      & 36.9               & 15.0    & 14.8             \\
mt       & 60.2      & 56.0               & 36.1    & 35.9    \\
tt       & 63.9   & 60.8      & 34.7          & 37.4             \\
fi       & 55.6      & 48.3               & 29.9    & 29.0              \\
sl       & 56.0   & 54.6      & 29.0          & 26.1   \\
pl       & 59.3      & 56.0               & 27.3    & 25.7      \\
\midrule
Avg & 51.1&	46.5&	23.9&22.2
\\
\bottomrule
\end{tabular}
\end{center}
\end{table}

In this section, we analyze the importance of pretraining, cross-lingual pretraining, lexicon construction strategies as well as the impact of different phonemizers. 
We use the CommonVoice benchmark for these experiments (Table~\ref{tab:data}).



\subsubsection{Effect of multilingual pretraining}
\begin{table}[t]
\caption{Effect of lexicon construction strategies (\textsection\ref{sec:mapping}) and different phonemizers (\textsection\ref{sec:preprocess}) on CommonVoice in terms of PER: tr2tgt denotes a lexicon constructed by mapping training language phonemes to target language phonemes and tgt2tr denotes the reverse strategy. 
Average PER excludes "eu" and "ia" since they are not supported by Phonetisaurus. \label{tab:cv_res_abl2}}
\begin{center}
\begin{tabular}{@{}ccccc@{}}
\toprule
 Phonemizer  & \multicolumn{2}{c}{Espeak}  & \multicolumn{2}{c}{Phonetisaurus}  \\
 \cmidrule(lr){2-3} \cmidrule(lr){4-5}
  Lexicon & tr2tgt      & tgt2tr     & tr2tgt      & tgt2tr   \\
\midrule
Avg & 24.5&	24.6&31.7&	32.4
\\
\bottomrule
\end{tabular}
\end{center}
\end{table}

Multilingual pretraining plays an important role for the model to perform well on unseen languages.
Table \ref{tab:cv_res_abl1} shows that accuracy without pretraining performs vastly less well than pretraining-based approaches, even when the amount of labeled data is increased by up to a factor of 20.
This is inline with prior work on automatic speech recognition~\cite{baevski2020wav2vec}.
Furthermore, multilingual pre-training (XLSR-53) performs better than monolingual pretraining on English data (w2v LV-60K) on every single language.

\subsubsection{Comparison of lexicon and phonemizers}
\label{sec:exp_lexicons}

Next, we compare the decoding performance with different lexicons.
Table~\ref{tab:cv_res_abl2} shows that tr2tgt is slightly better than tgt2tr on average for different phonemizers. 
Different phonemizers can generate fairly different phoneme sequences given the same word transcriptions which may impact the final performance of our models.
To better understand the impact of this, we use both Espeak and Phonetisaurus (\textsection\ref{sec:preprocess}) and evaluate them on both types of lexicon construction techniques.
Table~\ref{tab:cv_res_abl2} indicates that both phonemizers show the same trend in performance for tr2tgt/tgt2tr.

\section{Conculusion}

In this work, we investigate zero-shot transfer learning on cross-lingual phoneme recognition using a cross-lingually pretrained self-supervised model.
Pretraining vastly improves accuracy over no pretraining, even when a moderate amount of labeled data is used, and cross-lingual pretraining performs better than monolingual pretraining.
Our simple approach of fine-tuning a large pretrained model performs better than prior work which only used the feature extractor of a monolingually pre-trained wav2vec 2.0 model and which relied on task-specific architectures such as language embeddings.
We also show that our approach performs on par to the recently introduced unsupervised speech recognition work of~\cite{baevski2021unsupervised} which does not use labeled data from related languages and requires training separate models for each target language.


\vfill\pagebreak
\bibliographystyle{IEEEbib}
\small
\bibliography{refs}

\begin{thebibliography}{10}

\bibitem{dalmia2018sequence}
S.~Dalmia, R.~Sanabria, F.~Metze, and A.~Black,
\newblock ``Sequence-based multi-lingual low resource speech recognition,''
\newblock in {\em Proc. of ICASSP}. IEEE, 2018.

\bibitem{pratap2020massively}
V.~Pratap, A.~Sriram, et~al.,
\newblock ``Massively multilingual asr: 50 languages, 1 model, 1 billion
  parameters,''
\newblock {\em arXiv preprint arXiv:2007.03001}, 2020.

\bibitem{oord2018cpc}
A{\"{a}}. Oord, Y.~Li, and O.~Vinyals,
\newblock ``Representation learning with contrastive predictive coding,''
\newblock in {\em Proc. of NeurIPS}, 2018.

\bibitem{chung2018speech2vec}
Y.-A. Chung and J.~Glass,
\newblock ``Speech2vec: A sequence-to-sequence framework for learning word
  embeddings from speech,''
\newblock in {\em Proc. of Interspeech}, 2018.

\bibitem{chung2019unsupervised}
Y.-A. Chung, W.-N. Hsu, H.~Tang, and J.~Glass,
\newblock ``An unsupervised autoregressive model for speech representation
  learning,''
\newblock in {\em Proc. of Interspeech}, 2019.

\bibitem{baevski2020wav2vec}
A.~Baevski, H.~Zhou, A.~Mohamed, and M.~Auli,
\newblock ``wav2vec 2.0: A framework for self-supervised learning of speech
  representations,''
\newblock {\em arXiv preprint arXiv:2006.11477}, 2020.

\bibitem{hsu2021hubert}
W.-N. Hsu et~al.,
\newblock ``Hubert: Self-supervised speech representation learning by masked
  prediction of hidden units,''
\newblock {\em arXiv preprint arXiv:2106.07447}, 2021.

\bibitem{synnaeve2019end}
G.~Synnaeve, Q.~Xu, et~al.,
\newblock ``End-to-end {ASR}: from {Supervised} to {Semi}-{Supervised}
  {Learning} with {Modern} {Architectures},''
\newblock {\em arXiv}, vol. abs/1911.08460, 2019.

\bibitem{xu2020iterative}
Q.~Xu, T.~Likhomanenko, J.~Kahn, A.~Hannun, G.~Synnaeve, and R.~Collobert,
\newblock ``Iterative pseudo-labeling for speech recognition,''
\newblock {\em arXiv}, 2020.

\bibitem{likhomanenko2020slimipl}
T.~Likhomanenko, Q.~Xu, J.~Kahn, G.~Synnaeve, and R.~Collobert,
\newblock ``slimipl: Language-model-free iterative pseudo-labeling,''
\newblock {\em arXiv preprint arXiv:2010.11524}, 2020.

\bibitem{xu2021self}
Qiantong X., Alexei B., et~al.,
\newblock ``Self-training and pre-training are complementary for speech
  recognition,''
\newblock in {\em Proc. of ICASSP}. IEEE, 2021.

\bibitem{park2020improved}
D.~Park, Y.~Zhang, Y.~Jia, et~al.,
\newblock ``Improved noisy student training for automatic speech recognition,''
\newblock {\em arXiv preprint arXiv:2005.09629}, 2020.

\bibitem{liu2018completely}
D.~Liu, K.-Y. Chen, H.-Y. Lee, and L.~s.~Lee,
\newblock ``Completely unsupervised phoneme recognition by adversarially
  learning mapping relationships from audio embeddings,''
\newblock in {\em Proc. of Interspeech}, 2018.

\bibitem{chen2019completely}
K.-Y. Chen, C.-P. Tsai, D.-R. Liu, et~al.,
\newblock ``Completely unsupervised speech recognition by a generative
  adversarial network harmonized with iteratively refined hidden markov
  models,''
\newblock in {\em Proc. of Interspeech}, 2019.

\bibitem{baevski2021unsupervised}
A.~Baevski, W.-N. Hsu, A.~Conneau, and M.~Auli,
\newblock ``Unsupervised speech recognition,''
\newblock {\em arXiv preprint arXiv:2105.11084}, 2021.

\bibitem{gaozero}
H.~Gao, J.~Ni, Y.~Zhang, K.~Qian, et~al.,
\newblock ``Zero-shot cross-lingual phonetic recognition with external language
  embedding,''
\newblock in {\em Proc. of Interspeech}, 2021.

\bibitem{li2020universal}
X.~Li, S.~Dalmia, J.~Li, et~al.,
\newblock ``Universal phone recognition with a multilingual allophone system,''
\newblock in {\em Proc. of ICASSP}. IEEE, 2020.

\bibitem{jacobs2021multilingual}
C.~Jacobs and H.~Kamper,
\newblock ``Multilingual transfer of acoustic word embeddings improves when
  training on languages related to the target zero-resource language,''
\newblock {\em arXiv preprint arXiv:2106.12834}, 2021.

\bibitem{yan2021differentiable}
B.~Yan, S.~Dalmia, D.~Mortensen, F.~Metze, and S.~Watanabe,
\newblock ``Differentiable allophone graphs for language-universal speech
  recognition,''
\newblock {\em arXiv preprint arXiv:2107.11628}, 2021.

\bibitem{li2020towards}
X.~Li, S.~Dalmia, D.~Mortensen, et~al.,
\newblock ``Towards zero-shot learning for automatic phonemic transcription,''
\newblock in {\em Proc. of AAAI}, 2020.

\bibitem{winata2020adapt}
G.~Winata, G.~Wang, C.~Xiong, and S.~Hoi,
\newblock ``Adapt-and-adjust: Overcoming the long-tail problem of multilingual
  speech recognition,''
\newblock {\em arXiv preprint arXiv:2012.01687}, 2020.

\bibitem{schultz2002globalphone}
T.~Schultz,
\newblock ``Globalphone: a multilingual speech and text database developed at
  karlsruhe university,''
\newblock in {\em ICSLP}, 2002.

\bibitem{conneau2020unsupervised}
A.~Conneau, A.~Baevski, R.~Collobert, A.~Mohamed, and M.~Auli,
\newblock ``Unsupervised cross-lingual representation learning for speech
  recognition,''
\newblock {\em arXiv preprint arXiv:2006.13979}, 2020.

\bibitem{ardila2019common}
R.~Ardila et~al.,
\newblock ``Common voice: A massively-multilingual speech corpus,''
\newblock {\em arXiv preprint arXiv:1912.06670}, 2019.

\bibitem{gales2014speech}
M.~Gales, K.~M Knill, A.~Ragni, and S.~Rath,
\newblock ``Speech recognition and keyword spotting for low-resource languages:
  Babel project research at cued,''
\newblock in {\em SLTU}, 2014.

\bibitem{pratap2020mls}
V.~Pratap, Q.~Xu, A.~Sriram, G.~Synnaeve, and R.~Collobert,
\newblock ``Mls: A large-scale multilingual dataset for speech research,''
\newblock {\em arXiv preprint arXiv:2012.03411}, 2020.

\bibitem{devlin2018bert}
J.~Devlin, M.-W. Chang, K.~Lee, and K.~Toutanova,
\newblock ``Bert: Pre-training of deep bidirectional transformers for language
  understanding,''
\newblock {\em arXiv}, 2018.

\bibitem{baevski2019vqwav2vec}
A.~Baevski, S.~Schneider, and M.~Auli,
\newblock ``vq-wav2vec: Self-supervised learning of discrete speech
  representations,''
\newblock in {\em Proc. of ICLR}, 2020.

\bibitem{vaswani2017transformer}
A.~Vaswani, N.~Shazeer, N.~Parmar, J.~Uszkoreit, L.~Jones, and et~al.,
\newblock ``Attention is all you need,''
\newblock in {\em Proc. of NIPS}, 2017.

\bibitem{jegou2011ieee}
H.~Jegou, M.~Douze, and C.~Schmid,
\newblock ``Product quantization for nearest neighbor search,''
\newblock {\em IEEE Trans. Pattern Anal. Mach. Intell.}, 2011.

\bibitem{jang2016gumbel}
Eric Jang, Shixiang Gu, and Ben Poole,
\newblock ``Categorical reparameterization with gumbel-softmax,''
\newblock {\em arXiv}, 2016.

\bibitem{Mortensen-et-al:2016}
David~R. M., Patrick L., et~al.,
\newblock ``Panphon: {A} resource for mapping {IPA} segments to articulatory
  feature vectors,''
\newblock in {\em Proc. of COLING}. 2016, {ACL}.

\bibitem{hasegawa2020grapheme}
M.~Hasegawa-Johnson et~al.,
\newblock ``Grapheme-to-phoneme transduction for cross-language asr,''
\newblock in {\em SLSP}. Springer, 2020.

\bibitem{ott2019fairseq}
M.~Ott, S.~Edunov, A.~Baevski, et~al.,
\newblock ``fairseq: A fast, extensible toolkit for sequence modeling,''
\newblock {\em arXiv preprint arXiv:1904.01038}, 2019.

\bibitem{graves2006connectionist}
Alex Graves, Santiago Fern{\'a}ndez, Faustino Gomez, and J{\"u}rgen
  Schmidhuber,
\newblock ``Connectionist temporal classification: labelling unsegmented
  sequence data with recurrent neural networks,''
\newblock in {\em Proc. of ICML}, 2006.

\bibitem{pratap2019wav2letter++}
V.~Pratap, A.~Hannun, Q.~Xu, et~al.,
\newblock ``Wav2letter++: A fast open-source speech recognition system,''
\newblock in {\em Proc. of ICASSP}. IEEE, 2019.

\end{thebibliography}

\vfill\pagebreak
\newpage

\onecolumn

\begin{appendices}
\normalsize
\begin{table*}[h!]
\caption{Statistics of languages from CommonVoice and the ones that are supported in Espeak and Phonetisaurus phonemizers. The languages denoted with $*$ are potentially not well supported by Espeak phonemizer, so we manually removed them in either train or test set. \label{tab:data_cv}}
\begin{center}
\begin{tabular}{@{}cccccccccc@{}}
\toprule
\multirow{2}{*}{\textbf{Dataset}} & \multirow{2}{*}{\textbf{Code}} & \multirow{2}{*}{\textbf{Lang}} & \multirow{2}{*}{\textbf{Family}} & \multirow{2}{*}{\textbf{Split}} & \multicolumn{3}{c}{\textbf{Hours}} & \multicolumn{2}{c}{\textbf{Phonemizer}} \\
                                  &                                &                                &                                  & & train      & valid      & test     & Espeak    & Phonetisaurus               \\
\cmidrule(lr){1-1} \cmidrule(lr){2-5} \cmidrule(lr){6-8} \cmidrule(lr){9-10}   
\multirow{42}{*}{CV}              & eo                             & esperanto                      & Constructed      &train                & 34.0       & 13.3       & 14.3     & eo        &                             \\
                                  & lt                             & lithuanian                     & Baltic           &train                & 1.2        & 0.4        & 0.7      & lt        & lithuanian\_4\_2\_2.fst     \\
                                  & cy                             & welsh                          & Celtic           &train                & 9.1        & 6.8        & 7.0      & cy        &                             \\
                                  & ta                             & tamil                          & Dravidian        &train                & 2.4        & 2.2        & 2.3      & ta        & tamil\_2\_3\_3.fst          \\
                                  & sv-SE                          & swedish                        & North Germanic   &train                & 2.1        & 1.7        & 1.8      & sv-SE     & swedish\_4\_4\_4.fst        \\
                                  & de                             & german                         & West Germanic     &train               & 392.7      & 25.0       & 25.5     & de        & german\_download.fst        \\
                                  & en                             & english                        & West Germanic    &train                & 893.5      & 27.2       & 26.0     & en        & english\_4\_2\_2.fst        \\
                                  & as                             & assamese                       & Indic           &train                 & 0.4        & 0.2        & 0.2      & as        & assamese\_4\_2\_3.fst       \\
                                  & hi                             & hindi                          & Indic           &train                 & 0.2        & 0.2        & 0.2      & hi        & hindi\_4\_2\_2.fst          \\
                                  & or                             & oriya                          & Indic           &train                 & 0.6        & 0.2        & 0.2      & or        &                             \\
                                  & fa                             & persian                        & Iranian         &train                 & 7.7        & 6.5        & 7.2      & fa        & persian\_2\_2\_2.fst        \\
                                  & ja$*$                             & japanese                       & Japonic      &train                    & 0.9        & 0.8        & 0.9      & ja        & japanese\_4\_4\_4.fst       \\
                                  & id                             & indonesian                     & Austronesian    &train                 & 2.1        & 1.9        & 2.0      & id        & indonesian\_2\_4\_4.fst     \\
                                  & ca                             & catalan                        & Romance        &train                  & 441.5      & 24.0       & 24.9     & ca        &                             \\
                                  & es                             & spanish                        & Romance        &train                  & 235.1      & 25.0       & 25.7     & es        & spanish\_4\_3\_2.fst        \\
                                  & fr                             & french                         & Romance        &train                  & 424.5      & 24.0       & 25.1     & fr        & french\_8\_4\_3.fst         \\
                                  & pt                             & portuguese                     & Romance       &train                   & 7.8        & 5.6        & 6.1      & pt        & portuguese\_download.fst    \\
                                  & ar                             & arabic                         & Semitic       &train                   & 16.0       & 8.8        & 9.1      & ar        & arabic\_download.fst        \\
                                  & zh-CN                          & chinese                        & Sino-Tibetan   &train                  & 26.6       & 13.3       & 14.1     & cmn       & mandarin\_2\_4\_4.fst       \\
                                  & zh-TW                          & chinese                        & Sino-Tibetan   &train                  & 3.0        & 2.4        & 2.6      & cmn       & mandarin\_2\_4\_4.fst       \\
                                  & tr                             & turkish                        & Turkic         &train                  & 2.0        & 1.9        & 2.1      & tr        & turkish\_download.fst       \\
                                  & ky                             & kirghiz                        & Turkic         &train                  & 2.6        & 2.1        & 1.9      & ky        & kirghiz\_8\_2\_2.fst    \\
                                  & et                             & estonian                       & Uralic        &train                   & 5.5        & 4.7        & 4.6      & et        & estonian\_2\_4\_4.fst       \\
                                  & hu                             & hungarian                      & Uralic         &train                  & 4.3        & 1.7        & 1.9      & hu        & hungarian\_2\_4\_2.fst      \\
                                  & ru                             & russian                        & East Slavic     &train                 & 23.5       & 12.3       & 13.2     & ru        & russian\_download.fst       \\
                                  & cs                             & czech                          & West Slavic     &train                 & 7.3        & 5.0        & 5.0      & cs        & czech\_4\_4\_4.fst          \\
                                  & it                             & italian                        & Romance         &dev                 & 86.2       & 21.0       & 22.1     & it        & italian\_8\_2\_3.fst        \\
                                  & eu                             & basque                         & Language isolate     &test            & 10.9       & 7.8        & 8.2      & eu        &                             \\
                                  & ia                             & interlingua                    & Constructed       &test               & 2.2        & 1.5        & 0.8      & ia        &                             \\
                                  & lv                             & latvian                        & Baltic           &test                & 1.9        & 1.6        & 1.6      & lv        & latvian\_2\_4\_4.fst        \\
                                  & ka                             & georgian                       & South Caucasian   &test               & 1.6        & 0.9        & 1.0      & ka        & georgian\_4\_2\_3.fst       \\
                                  & nl                             & dutch                          & West Germanic    &test                & 11.5       & 6.4        & 7.0      & nl        & dutch\_download.fst         \\
                                  & el                             & greek                          & Hellenic      &test                   & 2.8        & 1.5        & 1.8      & el        & greek\_2\_2\_2.fst          \\
                                  & ro                             & romanian                       & Romance     &test                     & 3.6        & 1.0        & 2.0      & ro        & romanian\_2\_3\_3.fst       \\
                                  & mt                             & maltese                        & Semitic     &test                     & 2.3        & 1.8        & 2.1      & mt        & maltese\_2\_4\_4.fst        \\
                                  & tt                             & tatar                          & Turkic      &test                     & 11.5       & 2.0        & 4.4      & tt        & tatar\_2\_2\_2.fst          \\
                                  & fi                             & finnish                        & Uralic      &test                     & 0.5        & 0.5        & 0.6      & fi        & finnish\_2\_4\_4.fst        \\
                                  & sl                             & slovenian                      & South Slavic     &test                & 1.9        & 0.5        & 0.7      & sl        & slovenian\_2\_4\_4.fst      \\
                                  & pl                             & polish                         & West Slavic     &test                 & 9.3        & 6.6        & 7.0      & pl        & polish\_2\_2\_2.fst         \\
                                  & ga-IE$*$                         & irish                          & Celtic        &test                   & 0.5        & 0.4        & 0.5      & ga        &                             \\
                                  & zh-HK$*$                         & chinese                        & Sino-Tibetan   &test                  & 3.9        & 3.1        & 3.6      & yue       & yue\_2\_2\_4.fst            \\
                                  & pa-IN$*$                          & punjabi                        & Indic     &test                       & 0.2        & 0.1        & 0.1      & pa        & panjabi\_4\_4\_4.fst        \\
                                \bottomrule
\end{tabular}
\end{center}
\end{table*}
\begin{table*}[h!]
\caption{Statistics of languages from Babel and the ones that are supported in Espeak and Phonetisaurus phonemizers. \label{tab:data_babel}}
\begin{center}
\begin{tabular}{@{}ccccccccc@{}}
\toprule
\multirow{2}{*}{\textbf{Dataset}} & \multirow{2}{*}{\textbf{Code}} & \multirow{2}{*}{\textbf{Lang}} & \multirow{2}{*}{\textbf{Family}} & \multicolumn{3}{c}{\textbf{Hours}} & \multicolumn{2}{c}{\textbf{Phonemizer}} \\
                                  &                                &                                &                                  & train      & valid      & test     & Espeak     & Phonetisaurus              \\
\cmidrule(lr){1-1} \cmidrule(lr){2-4} \cmidrule(lr){5-7} \cmidrule(lr){8-9} 
\multirow{21}{*}{Babel}           & 307                            & Amharic                        & Semitic                          & 39.4       & 4.4        & 11.7     & am         & amharic\_8\_2\_4.fst       \\
                                  & 103                            & Bengali                        & Indic                            & 56.4       & 6.3        & 10.0     & bn         & bengali\_4\_3\_2.fst       \\
                                  & 301                            & Cebuano                        &                                  & 37.4       & 4.2        & 10.4     &            & cebuano\_4\_3\_2.fst       \\
                                  & 201                            & Haitian                        & Creole                           & 61.0       & 6.7        & 10.8     & ht         & haitian\_8\_3\_3.fst       \\
                                  & 402                            & Javanese                       & Austronesian                     & 41.1       & 4.6        & 11.4     &            & javanese\_4\_2\_2.fst      \\
                                  & 202                            & Swahili                        & Bantu                            & 40.1       & 4.5        & 10.7     & sw         & swahili\_4\_2\_2.fst       \\
                                  & 204                            & Tamil                          & Dravidian                        & 62.6       & 7.0        & 11.6     & ta         & tamil\_2\_3\_3.fst         \\
                                  & 107                            & Vietnamese                     & Austroasiatic                    & 78.8       & 8.8        & 11.0     & vi         & vietnamese\_2\_2\_2.fst    \\
                                  & 102                            & Assamese                       & Indic                            & 54.8       & 6.1        & 10.0     & as         & assamese\_4\_2\_3.fst      \\
                                  & 403                            & Dholuo                         &                                  & 37.6       & 4.1        & 10.1     &            & luo\_4\_2\_2.fst           \\
                                  & 305                            & Guarani                        & South American Indian            & 38.9       & 4.3        & 10.6     & gn         & guarani\_4\_2\_2.fst       \\
                                  & 306                            & Igbo                           & Niger–Congo                      & 39.7       & 4.4        & 10.9     &            & igbo\_2\_3\_4.fst          \\
                                  & 302                            & Kazakh                         & Turkic                           & 36.1       & 4.0        & 9.8      & kk         & kazakh\_2\_3\_2.fst        \\
                                  & 104                            & Pashto                         & Indo-European                    & 70.7       & 7.8        & 10.0     &            & pushto\_8\_3\_2.fst        \\
                                  & 106                            & Tagalog                        & Austronesian                     & 76.2       & 8.6        & 10.7     &            & tagalog\_4\_2\_3.fst       \\
                                  & 303                            & Telugu                         & Dravidian                        & 38.1       & 4.3        & 9.9      & te         & telugu\_4\_4\_4.fst        \\
                                  & 105                            & Turkish                        & Turkic                           & 70.0       & 7.8        & 9.9      & tr         & turkish\_download.fst      \\
                                  & 206                            & Zulu                           & Niger–Congo                      & 56.4       & 6.2        & 10.5     &            & zulu\_4\_4\_3.fst          \\
                                  & 404                            & Georgian                       & South Caucasian                  & 45.5       & 5.1        & 12.4     & ka         & georgian\_4\_2\_3.fst      \\
                                  & 101                            & Cantonese                      & Sino-Tibetan                     & 120.3      & 13.5       & 17.0     & yue        & yue\_2\_2\_4.fst           \\
                                  & 203                            & Lao                            & Tai–Kadai                        & 59.2       & 6.5        & 10.6     &            & lao\_2\_2\_2.fst          \\
\bottomrule
\end{tabular}
\end{center}
\end{table*}
\begin{table*}[h!]
\caption{Comparison of PER on the test set of a subset of Common Voice languages. tr2tgt lexicon is used in beam-search decoding by default, while tgt2tr lexicon is used only for the columns denoted with $*$. The numbers in the parenthesis next to each pre-trained model is the maximum number of hours per language in the training set. \label{tab:cv_res}
}
\begin{center}
\setlength{\tabcolsep}{4.5pt}
\begin{tabular}{@{}ccccccccccccc@{}}
\toprule
 Pretrain  & \multicolumn{2}{c}{- (10)} & \multicolumn{2}{c}{- (200)} & \multicolumn{2}{c}{EN - LV (10)} & \multicolumn{3}{c}{XLSR - 53 (10)}      & \multicolumn{3}{c}{XLSR - 53 (10)} \\
 Phonemizer  & \multicolumn{2}{c}{Espeak} & \multicolumn{2}{c}{Espeak}  & \multicolumn{2}{c}{Espeak}       & \multicolumn{3}{c}{Espeak}              & \multicolumn{3}{c}{Phonetisaurus}  \\
   & viterbi      & n-gram      & viterbi       & n-gram      & viterbi         & n-gram         & viterbi & n-gram        & n-gram*       & viterbi    & n-gram    & n-gram*   \\
\cmidrule(lr){2-3} \cmidrule(lr){4-5} \cmidrule(lr){6-7} \cmidrule(lr){8-10} \cmidrule(lr){11-13}
it & 56.6         & 47.5        & 50.1          & 41.8        & 31.8            & 16.9           & 26.0    & \textbf{13.9} & 14.3          & 26.6       & 18.1      & 17.8      \\
eu & 51.2         & 45.6        & 39.7          & 32.1        & 24.8            & 16.3           & 20.8    & 13.7          & \textbf{12.2} & -          & -         & -         \\
ia & 38.9         & 27.8        & 30.9          & 23.0        & 12.7            & 6.7            & 10.7    & 6.1           & \textbf{6.0}  & -          & -         & -         \\
lv & 65.2         & 59.8        & 62.7          & 56.5        & 41.9            & 33.5           & 39.9    & \textbf{32.3} & 34.0          & 50.0       & 40.5      & 62.4      \\
ka & 61.8         & 56.1        & 54.6          & 48.9        & 29.1            & 24.0           & 30.5    & \textbf{23.8} & 24.3          & 34.7       & 26.6      & 25.5      \\
nl & 66.3         & 56.8        & 63.0          & 56.1        & 46.8            & 30.5           & 37.1    & \textbf{19.8} & 22.7          & 37.3       & 24.4      & 27.8      \\
el & 49.5         & 40.6        & 42.1          & 33.7        & 18.9            & 10.7           & 17.3    & 10.4          & \textbf{9.9}  & 36.2       & 32.2      & 22.9      \\
ro & 45.7         & 34.7        & 46.7          & 36.9        & 21.3            & 15.0           & 20.1    & 14.8          & \textbf{12.6} & 28.5       & 16.2      & 17.5      \\
mt & 66.3         & 60.2        & 62.0          & 56.0        & 47.4            & 36.1           & 46.6    & \textbf{35.9} & 36.1          & 43.3       & 34.7      & 37.5      \\
tt & 68.2         & 63.9        & 65.1          & 60.8        & 46.2            & 34.7           & 48.4    & 37.4          & \textbf{35.5} & 49.1       & 45.6      & 35.3      \\
fi & 58.8         & 55.6        & 53.8          & 48.3        & 36.8            & 29.9           & 36.8    & 29.0          & \textbf{27.1} & 43.1       & 34.0      & 37.2      \\
sl & 62.9         & 56.0        & 60.5          & 54.6        & 43.5            & 29.0           & 40.6    & \textbf{26.1} & 27.4          & 33.4       & 26.1      & 23.9      \\
pl & 62.3         & 59.3        & 60.2          & 56.0        & 36.1            & 27.3           & 32.8    & \textbf{25.7} & 27.1          & 51.8       & 49.9      & 48.1       \\
\midrule
Avg & 58.0 &	51.1&	53.2&	46.5&	33.6&	23.9&	31.4&	\textbf{22.2}&	22.3&	39.5&	31.7&	32.4
\\
\bottomrule
\end{tabular}
\end{center}
\end{table*}
\twocolumn

\section{Dataset Details}

In this section, we summarize the details of CommonVoice and BABEL datsets. Specifically we list the code and name of each language together with the family they belong to. We also show the duration in hours of each split of each language. The amount of training data varies a lot in CommonVoice dataset. 
We subsample the training data for high resource languages to avoid bias. Additionally, for each language, we also provide the specific language identifier we used in Espeak and the specific finite state transducers\footnote{https://github.com/uiuc-sst/g2ps} in Phonetisaurus. We can see that Espeak covers more languages in CommonVoice, while Phonetisaurus covers more languages in BABEL. 

The languages in Table \ref{tab:data_cv} are ordered first by splits (training, validation and test) and then they are grouped by families.


\begin{figure}[t]
  \centering
  \includegraphics[width=0.54\textwidth]{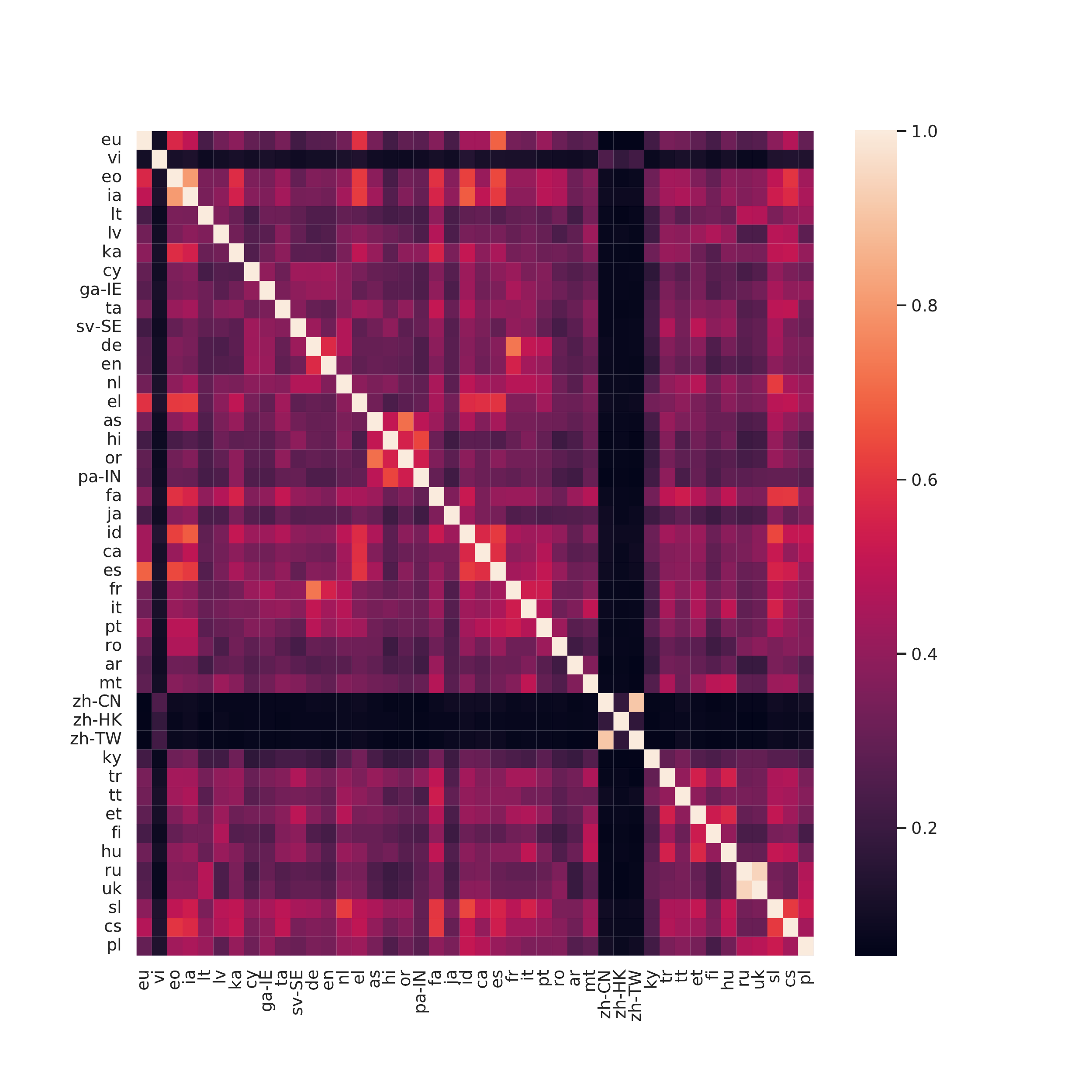}
  \caption{
      Correlation between each pair of languages in CommonVoice dataset. \label{fig:cor}
  }
\end{figure}

\section{Language correlation}

We simply denote the correlation between each pair of languages by $cor(l_1, l_2) = \frac{|vocab(l_1)~\cup~vocab(l_2)|}{|vocab(l_1)~\cup~vocab(l_2)|}$, where $l_1$ and $l_2$ are two languages and $vocab(\cdot)$ denotes the phoneme vocabulary of a given language. Figure \ref{fig:cor} shows the correlations between pairs of CommonVoice languages. Since languages are ordered purely by family, it is reasonable to see high correlations on the diagonal blocks. However, this high correlation also scatter around the whole plot, meaning that IPA phoneme symbols are commonly shared across different languages and it is good for zero-shot transfer learning. Besides, Vietnamese (vi) and Chinese faimily (zh-CN, zh-TW, zh-HK) seems isolated to others, as their phoneme symbols include tones. Specifically, vowels like 'ou' can be denoted as one of the following: 'ou1', 'ou2', 'ou3', 'ou4', ou5' and 'ou6'. They are intrinsically both hard to learn and hard to predict.

\section{Full comparison on CommonVoice}

We summarize all the results on CommonVoice in Table \ref{tab:cv_res}. Apart from the analysis in the ablation section, we can also find that beam-search decoding consistently helps to improve the model performance for all languages in all the settings. The results in Table \ref{tab:cv_res_abl2} shows that the trend of accuracy on unseen languages is similar across phonemizers. 

\section{Full results on BABEL}

\begin{table}[t]
\begin{center}
\caption{Comparison to prior zero-shot work~\cite{gaozero} in terms of phonetic token error rate (PTER) on the test sets of a subset of BABEL languages. Cantonese and Lao are the unseen languages. BB and CV represents BABEL and CommonVoice dataset and the following numbers are the number of the languages included in the training set.\label{tab:gao_res_full}
}
\setlength{\tabcolsep}{3pt}
\begin{tabular}{@{}cccccc@{}}
\toprule
BB Data          & BB-6\cite{gaozero} & BB-6 & BB-19 & - & BB-19 \\
Other Data       & CGN+GP & - & - & CV-21 & CV-21 \\
\# hours / lang &        all              &       all           &        all           & 10             & 10                        \\
\# hours total               & 1,492                & 317              & 935               & 118            & 298                       \\
\midrule
Bengali                   & 38.2                 & \textbf{36.1}    & 35.4              & 53.2           & 40.7                      \\
Vietnamese                & \textbf{32.0}        & 40.7             & 42.1              & 71.0           & 63.3                      \\
Zulu                      & 35.2                 & \textbf{34.6}    & 34.8              & 61.0           & 44.1                      \\
Amharic                   & 38.0                 & \textbf{35.5}    & 35.5              & 63.2           & 42.8                      \\
Javanese                  & 44.2                 & \textbf{40.2}    & 40.8              & 57.4           & 49.1                      \\
Georgian                  & 38.6                 & \textbf{27.6}    & 43.8              & 51.6           & 43.2                      \\
\midrule
Cantonese                 & 73.1                 & 73.6             & 72.6              & 70.9           & \textbf{63.6}             \\
Lao                       & 69.3                 & 70.3             & 70.2              & 72.1           & \textbf{63.7}    \\ 
\bottomrule         
\end{tabular}

\end{center}
\end{table}
As shown in Table \ref{tab:gao_res_full}, with finetuning on only the BABEL subset of \cite{gaozero}'s training data, our method performs better on the supervised languages already, indicating that the wav2vec Transformer blocks, that are not included in \cite{gaozero}, benefit the model learning a lot. Additionally, models trained with mixed CommonVoice and BABEL data generalize better than the ones trained on either one of them on the unseen languages. It also surpasses \cite{gaozero} with using an extra learned language encoder.

\end{appendices}
\end{document}